\documentclass[12pt,letterpaper]{article}

\usepackage{textcomp}
\usepackage{fullpage}
\usepackage{amsfonts}
\usepackage{verbatim}
\usepackage[english]{babel}
\usepackage{pifont}
\usepackage{color}
\usepackage{setspace}
\usepackage{indentfirst}
\usepackage[normalem]{ulem}
\usepackage{booktabs}
\usepackage{float}
\usepackage{latexsym}
\usepackage{url}
\usepackage{epsfig}
\usepackage{graphicx}
\usepackage{amssymb}
\usepackage{amsmath}
\usepackage{bm}
\usepackage{array}
\usepackage[version=4]{mhchem}
\usepackage{ifthen}
\usepackage{caption}
\usepackage{hyperref}
\usepackage{amsthm}
\usepackage{amstext}
\usepackage{enumerate}
\usepackage{csquotes}
\usepackage{setspace}
\usepackage{longtable}
\usepackage{lscape}
\usepackage{ragged2e} 

\raggedright
\renewcommand{\section}[1]{%
\bigskip
\begin{center}
\begin{Large}
\normalfont\scshape #1
\medskip
\end{Large}
\end{center}}

\usepackage{lineno}

\begin{document}
\bigskip
\noindent 

\bigskip
\medskip
\begin{center}

\noindent{\Large \bf Prior choice affects ability of Bayesian neural networks to identify unknowns}

\bigskip

\noindent {\normalsize 
\textsc{Daniele Silvestro}$^{1,2}$ and
\textsc{Tobias Andermann}$^{3,4}$
\noindent {\small \it 

$^1$Department of Biology, University of Fribourg, Switzerland;\\
$^2$Swiss Institute of Bioinformatics, Lausanne, Switzerland; \\
$^3$Department of Biological and Environmental Sciences, University of Gothenburg, Sweden;\\
$^4$Gothenburg Global Biodiversity Center, Sweden;\\
}
}
\medskip
daniele.silvestro[at]unifr.ch, tobias.andermann[at]bioenv.gu.se\\

\end{center}
\medskip

\vspace{1.5in}
\justify


\subsubsection{Abstract}
Deep Bayesian neural networks (BNNs) are a powerful tool, though computationally demanding, to perform parameter estimation while jointly estimating uncertainty around predictions. 
BNNs are typically implemented using arbitrary normal-distributed prior distributions on the model parameters.
Here, we explore the effects of different prior distributions on classification tasks in BNNs and evaluate the evidence supporting the predictions based on posterior probabilities approximated by Markov Chain Monte Carlo sampling and by computing Bayes factors. 
We show that the choice of priors has a substantial impact on the ability of the model to confidently assign data to the correct class (true positive rates). 
Prior choice also affects significantly the ability of a BNN to identify out-of-distribution instances as unknown (false positive rates). 
When comparing our results against neural networks (NN) with Monte Carlo dropout we found that BNNs generally outperform NNs.
Finally, in our tests we did not find a single best choice as prior distribution. Instead, each dataset yielded the best results under a different prior, indicating that testing alternative options can improve the performance of BNNs.

\clearpage
\newpage

\section{Introduction}

Neural networks (NN) have become a widespread and increasingly powerful tool for classification and regressions,  with applications in a wide range of research fields including medicine \cite{wang2020}, biology \cite{zhou2015, Silvestro2019}, sociology \cite{severyn2015}, and finance \cite{bao2017}.
An NN maps a set of features (input layer) onto an output layer of desired size, e.g. the number of classes in a classification task, through one or multiple hidden layers \cite{Goodfellow2016}. 

While a lot of the development in machine learning classifiers has focused on improving the accuracy of the models and scalability of the algorithms \cite{wang2019deep, pytorch2019}, there is an increasing awareness of the tendency of neural networks to provide highly confident predictions, even when these are wrong \cite{Goodfellow2016, hendrycks2016}.  
Erroneous classifications with high confidence (false positives) reduce the reliability of a classifier and can even be harmful in some contexts, prompting increased discussion about artificial intelligence safety \cite{amodei2016}.
This is an evident problem when the neural network is applied to out-of-distribution data, i.e. data from a class that was not included in the training set \cite{hendrycks2016}. In this case, samples that would ideally be identified as \textit{unknown}, often result in highly confident assignment to a class, particularly when interpreting directly the probability of a predicted label from the output layer as a measure of confidence \cite{Gal2016}.

Several methods have been developed to quantify uncertainties in predictions using neural networks, for instance using Monte Carlo (MC) dropout \cite{Gal2016} and deep ensembles \cite{lakshminarayanan2016}. 
An explicit probabilistic approach to estimate uncertainties in the form of posterior credible intervals is the use of Bayesian neural networks (BNN; \cite{neal2012}), which place prior distributions (generally normal densities) on the weight parameters of an NN and sample them from their posterior distribution or an approximation of it. 
BNNs typically implement Markov chain Monte Carlo (MCMC) algorithms \cite{neal2012, hoffman2014nuts,wenzel2020} or variational inference \cite{graves2011, blundell2015} to approximate the posterior distributions of the parameters from which predictions are sampled proportionally to their posterior probability.

Because the parameters of an NN do not usually have a direct interpretation, it is difficult to define prior distributions reflecting prior knowledge about the weights. 
While some studies have described the theoretical implications of using different priors in BNNs
\cite{lee2004, vladimirova2018}, the sensitivity of BNN estimates to these choices remains poorly understood.

Here, we explore the effects of using different prior distributions on the performance of BNNs in classification tasks. 
We assess the accuracy and true positive rate for in-distribution data and false positive rate for out-of-distribution data under different priors. 
We evaluate the statistical support for predictions in two ways:
1) using posterior probabilities for each class as approximated by MCMC sampling and
2) by calculating Bayes factors to account for potentially uneven prior expectations across the classes.
We compare accuracy and true and false positive rates obtained from BNNs with analogous metrics obtained from a standard NN using MC dropout to estimate uncertainties.


\section{Methods}
\subsection{Models and implementation}

We implemented a BNN framework using Python v.3 based on the modules Numpy and Scipy and used a Metropolis Hastings Markov chain Monte Carlo (MCMC) algorithm to sample the weights from their posterior distribution. 
In our simulations we used fully connected deep neural networks \cite{lecun2015deeplearning} with two hidden layers varying the number on nodes for each dataset (see below), and with an additional bias node in the input layer. 
We used a rectified linear units function ($ReLU$) activation function for hidden layers \cite{ReLU2010} and
a softmax function in the output layer to obtain the parameters of the categorical distribution used to compute the likelihood of the data.
The BNN implementation used in this study is available here: \href{https://github.com/dsilvestro/npBNN}{github.com/dsilvestro/npBNN}.

\begin{figure}[h!]
\centering
\includegraphics[width=0.8\textwidth]{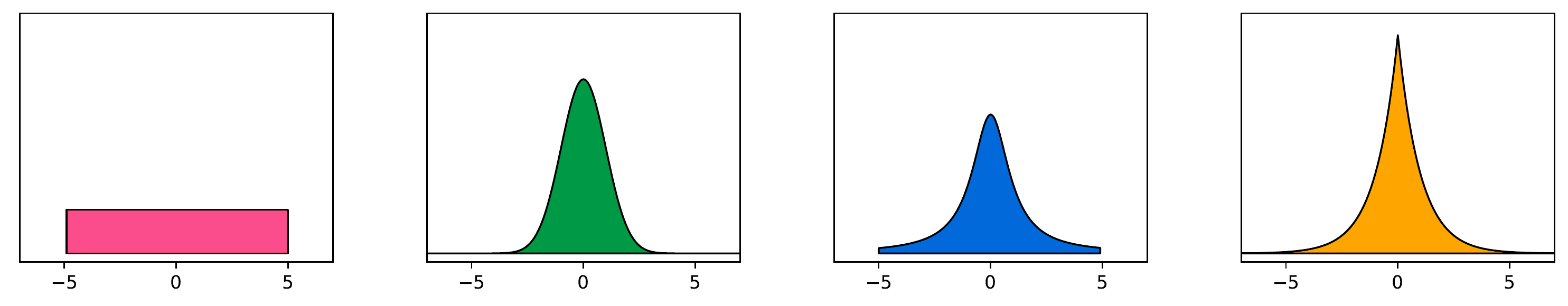}
\caption{Prior densities on weight distributions used in this study to assess their effect on the performance of BNNs.}
\label{figPriorPDF}
\end{figure}

We tested different prior distributions (Fig. \ref{figPriorPDF}) on the weights to explore their effect on the performance of posterior predictions:
\begin{itemize}
    \item Uniform: $P(w) \sim \mathcal{U}(-b,b)$ 
    \item Standard normal: $P(w) \sim \mathcal{N}(0,1)$
    \item Truncated Cauchy:  $ P(w) \sim  \begin{cases} \mathcal{C}(0,1), & \mbox{if } -b \geq w \leq b \\ 0, & \mbox{otherwise  } \end{cases} $
    \item Laplace:  $ P(w) \sim  \mathcal{L}(0, 1)$
 \end{itemize}
where the parameter $b$ defined the boundaries of the uniform distribution and of the truncated Cauchy distribution, in our analyses set to $b = 5$. 
Although the Cauchy distribution, unlike the uniform, does not need hard boundaries to be a proper prior, we found that the truncated Cauchy did not change noticeably the results in our analyses in terms of accuracy, but improved the converge of the MCMC, compared to a non truncated Cauchy prior. 

 While uniform priors result in the posterior distribution matching the likelihood surface (within the allowed range of model parameters), normal, Cauchy, and Laplace distributions introduce some shrinkage toward weights around 0. 
 However, their regularization effects differ in how they pull the parameters toward 0 by a constant proportion (normal), by constant amount (Laplace), or shrinking more strongly parameter values close 0 (Cauchy) \cite{Gelman_et_al2013}.

\subsection*{Datasets}
We tested the effects of different priors on BNNs based on three datasets. For each dataset, we excluded one or more classes from the training set and used them as out-of-distribution data to quantify the rate of false positives. 
All the data and scrips used to run and summarize the analyses are available here \href{https://doi.org/10.5281/zenodo.3816927}{\underline{doi:10.5281/zenodo.3816928}}.

I. Wine data --
To test the effect of prior choice on a simple classification task, we used the scikit-learn wine data (\href{https://scikit-learn.org/stable/}{scikit-learn.org}). 
The dataset consist of 178 samples with 13 numeric features, which result from a chemical analysis of Italian wines. The wines are classified into three classes based on their provenance. We rescaled all features using a min-max scaler prior to training.

II. Virus data --
As a possible real-world application, we compiled and analyzed a dataset of influenza virus RNA sequences,
obtained from the Influenza Research Database \cite{zhang2017influenza} (\href{www.fludb.org}{www.fludb.org}). 
Even though Influenza viruses are RNA viruses, the sequences that are available for download are coded in common DNA notation, i.e. the RNA nucleotide uracil (U) is coded as thymine (T). 
Our dataset was restricted to Influenza A viruses (e.g. the swine flu virus H1N1), which are categorized into subtypes based on the combination of two surface proteins hemagglutinin (H) and neuraminidase (N). 
We downloaded the coding RNA sequences (genes) of both of these proteins as sequence alignments (individually for each protein) and trimmed the alignments at the first methionine (Met) codon.
To deal with individual sequences of differing lengths, we trimmed the end of the alignments to ensure \textgreater 50\% of sequence coverage throughout the complete alignment. 
Next, we randomly selected 600 sequences for each subtype, only choosing samples with sequence data for both genes; the final dataset included 6,600 samples belonging to 11 Influenza A subtypes.
We then transformed the nucleotide sequences into numerical arrays, by determining the frequencies of each possible nucleotide triplet (i.e. all possible 3-letter permutations of the 4 nucleotides A,C,G,T), independently of the amino acid reading frame.
The resulting arrays of $4^{3} = 64$ frequencies of nucleotide triplets for each of the two segments were concatenated between the two genes, resulting in a set of 128 features for each instance.

III. Synthetic data --
We assessed the effect of prior choice on a classification task for which we expected a relatively low accuracy for in-distribution samples. 
To this end we simulated a dataset in which features were largely overlapping among classes. The dataset included 20 classes each represented by 199 instances. For each class $k$ we sampled 10 features drawn from beta distributions, such that the $i^{\text{th}}$ feature was a random draw from $\mathcal{B}(a_{k_i}, b_{k_i})$, where $a_{k_i}, b_{k_i} \sim \mathcal{U}(0.2,5)$.
Although each class is defined by a distinct set of shape parameters of the beta distribution, these are not guaranteed to be substantially different, as they are drawn from the same uniform distribution. As a consequence of this and the limited number of features, the dataset is expected to produce a classification accuracy $<$ 1 for in-distribution instances and potentially a high frequency of false positives for out-of-distribution data.

IV. MNIST data --
Finally we used a subset of the MNIST dataset of hand-written digits (10 classes) \cite{lecun1998mnist}, which includes 784 features. We randomly sampled 1,000 instances out of the original 60,000 to reduce computing time, even though this results in lower accuracy for in-distribution data.

\subsection{Training}
Since one of the aim of this study is to assess the ability of a BNN to distinguish between in-distribution and out-of-distribution data under different priors, we left out one or more classes during training and used them as test sets to estimate the false positive rates. 
We performed cross-validation by repeating this procedure 10 times, for datasets II--IV, while leaving out each time a randoms subset of classes.
We repeated all the analyses under the four prior distributions described above (Fig. \ref{figPriorPDF}). 
For all datasets we approximated the posterior distribution of the model parameters through 100 million MCMC iterations and assessed their convergence based on the effective sample sizes of the posterior samples.

For the wine dataset we split the data into in-distribution data (samples belonging to classes 1 and 2) and out-of-distribution data (samples belonging to class 3), thus training the BNN on two of the three classes. We set the number of nodes to 10 + 1 bias node and 5 for the first and second hidden layer, respectively.

For the virus dataset we generated 10 subsets of the training data, each containing the samples of 5 randomly selected subtypes (classes). The reason for this repeated random sampling of training classes is that some virus subtypes are more similar to each other than to others, and therefore the choice of the training classes is expected to affect the classification of out-of-distribution samples. 
Of the 600 instances for each class, we used 450 for training and 50 to monitor the test accuracy during the MCMC sampling,
while setting aside the remaining 100 instances as test set for in- and out-of-distribution predictions.
The BNN included 20 nodes + 1 bias node in the first hidden layer and 5 nodes in the second.

Of the 199 samples for each class included in dataset III, 99 were used for training and 100 as test set.
Across the 10 cross-validation replicates, we partitioned the synthetic data into two subset of 10 classes each and use them as in-distribution and out-of-distribution datasets. We configured the BNN with 15 nodes + 1 bias node in the first hidden layer and 10 in the second.

For the subset of the MNIST dataset, we trained the BNN using 500 instances from 5 classes randomly selected at each cross-validation replicate. The BNN included 5 nodes + 1 bias node in the first hidden layer and 5 in the second.
To evaluate the performance of the BNN we used the full MNIST test set (10,000 instances) split into in-distribution and out-of-distribution classes.

For comparison we used the same data to train a standard NN as implemented in Tensorflow v.2.1 (\href{https://tensorflow.org}{tensorflow.org}), using the same configurations 
implemented in the BNN analyses. We split the datasets into training (90\%, of which 20\% was used for validation during training), and test (10\%) sets and used the ADAM optimizer \cite{kingma2014adam} with default learning rate to minimize the cross-entropy loss function. The optimal number of epochs was selected as to minimize the validation cross-entropy loss.


\subsection{Predictions}
We sampled 5,000 posterior samples of the weights estimated by the BNN to perform predictions based on in-distribution and out-of-distribution test sets. 
We computed the following statistics to quantify the performance of the BNNs using different priors and that of NNs with MC dropout:
\begin{itemize}
    \item Accuracy (for in-distribution data): rate of correct predictions based on the class with highest posterior probability for BNNs and MC dropout support for NNs
    \item True positives (for in-distribution data): rate of correct predictions with significant statistical support quantified using posterior probability and Bayes Factors for BNN and MC dropout support for NNs (see below)
    \item False positives (for both in-distribution and out-of-distribution data): rate of erroneous predictions that received significant statistical support (based on the same thresholds applied for true positives).
\end{itemize}

We quantified the statistical support for BNN predictions in two ways. 
First we considered a predicted class as significantly supported if it was sampled by the MCMC with posterior probability (PP) greater than 0.95.
As an additional measure of evidence supporting a prediction we computed Bayes factors (BF) based on a comparison between posterior and prior sampling frequencies obtained by MCMC. BF is defined as the ratio between the posterior odds and the prior odds \cite{Kass_1995}. Thus for a class $k$, we computed
\begin{equation}
    BF = \frac{P(k|D_i)}{1 - P(k|D_i)} / \frac{P(k)}{1-P(k)} 
\end{equation}
where $P(k|D_i)$ is the posterior probability of the class, approximated for sample $D_i$ by the MCMC sampling frequency. 
We computed prior probability $P(k)$ empirically, as the sampling frequency of class $k$ for sample $D_i$ obtained from a distribution of BNN parameters sampled from the prior. This was achieved through an MCMC in which the acceptance probability of a state was only based on prior ratio, regardless of the likelihood ratio.
We used a threshold of $BF > 150$ to determine very strong support for a prediction \cite{Kass_1995}. 

To rank the performance of different BNN models based on in-distribution data we evaluated the balance between true and false positive rates. 
We computed the informedness as the difference between the true positive rate and the false positive rate \cite{powers2011}. The informedness reaches a value of 1 when both sensitivity and specificity are 1.

For standard NNs, we evaluated the statistical support of the prediction using the MC dropout technique, since a direct interpretation of the softmax probabilities is known to produce over-confident interpretations \cite{Gal2016} (but see \cite{hendrycks2016}).
We applied a dropout rate adjusted to the different datasets, which was set to 0.2 for the datasets I and II, 0.01 for dataset III, and 0.05 for dataset IV. Dropout rates were adjusted to best balance the true and false positive rates. 
For each instance we sampled 1,000 predictions and used a threshold frequency of 0.95 to determine if the predicted class received significant statistical support.

\section{Results}

\subsection{Prior choice and in-distribution data}
\begin{figure}[h!]
\centering
\includegraphics[width=0.9\textwidth]{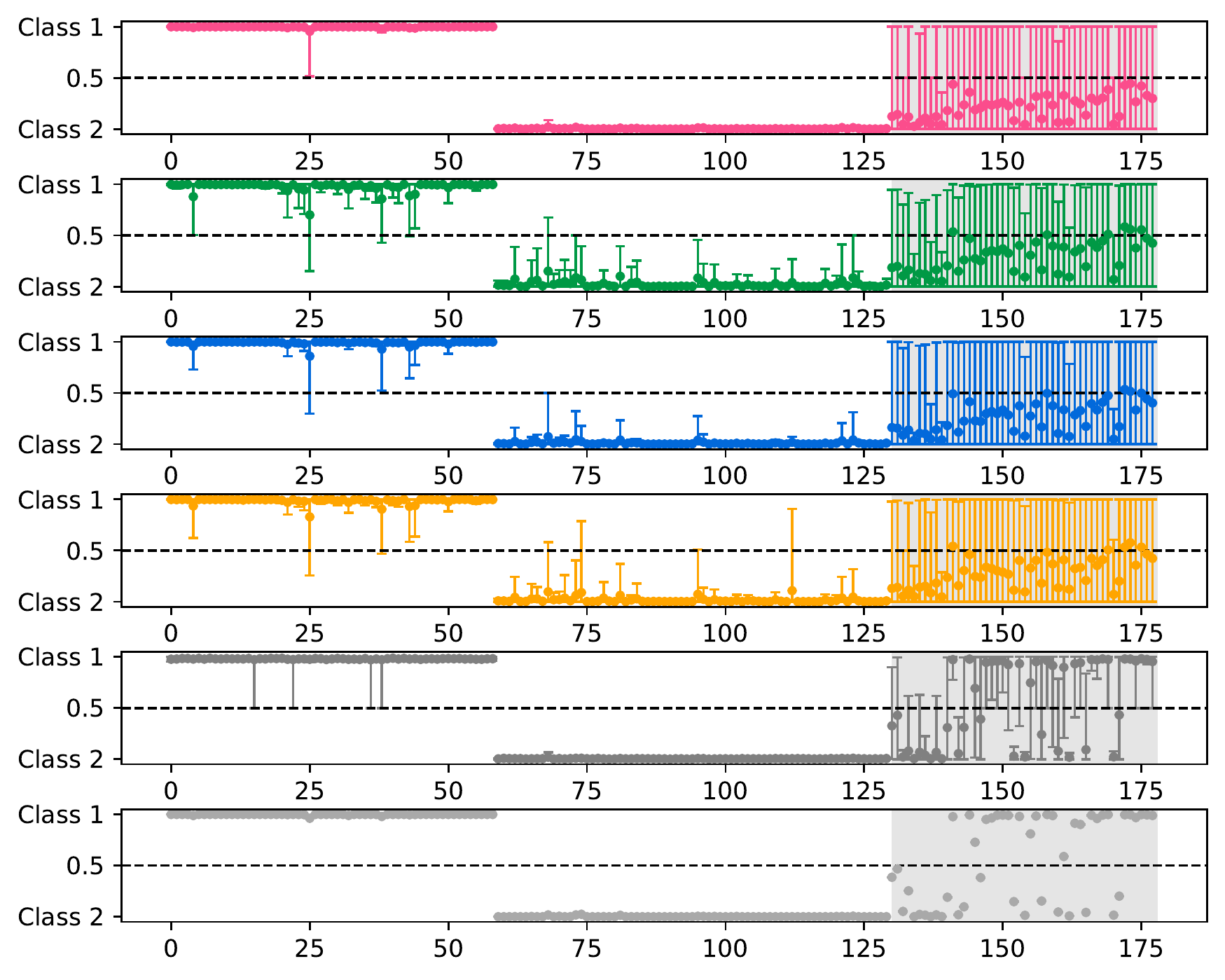}
\caption{Class probabilities (y-axis) predicted for dataset I (x-axis) using four BNN configurations (with uniform, normal, truncated-Cauchy, and Laplace priors, color-coded as in Fig. \ref{figPriorPDF}) as well as by an NN with MC dropout (dark gray) and a regular NN in light gray. All models were trained on data belonging to classes 1 and 2. Dots and whiskers show the mean and 95\% interval of label probabilities for each sample. For regular NNs, we report the probability of a class resulting from the SoftMax function at the output layer. Gray shaded areas indicate out-of-distribution instances belonging to the third class. 
}
\label{wine_data_predictions}
\end{figure}

\begin{figure}[h!]
\centering
\includegraphics[width=1\textwidth]{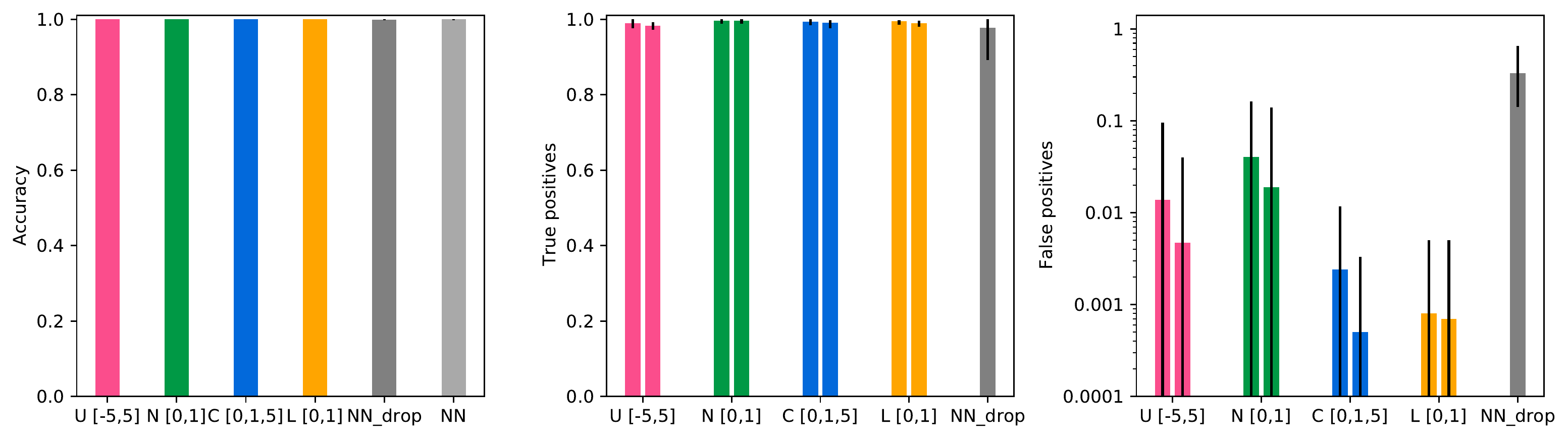}
\caption{Summary of BNN and NN predictions for dataset II, showing the test accuracy of the predictions (1st panel), the true positive rates (2nd panel), and false positive rates for out-of-distribution data (3rd panel). The results are shown for BNNs with four priors (Uniform, Normal, truncated-Cauchy, and Laplace (color-coded as in Fig. \ref{figPriorPDF}) and an NN with MC dropout (dropout rate = 0.2). The first panel additionally shows the prediction accuracy of a regular NN (no dropout, light grey).
True an false positives for BNNs were computed based on posterior probabilities (PP $>$ 0.95; bars on the left).
For each model we summarized the results of 10 independent runs, trained on a different random subset of 5 of the total 11 virus subtypes, and the whiskers show the range of values derived from these 10 subsets.
Note that the third panel is plotted in log-space for better visibility of small values.
 }
\label{virus_data_stats}
\end{figure}

\begin{figure}[h!]
\centering
\includegraphics[width=1\textwidth]{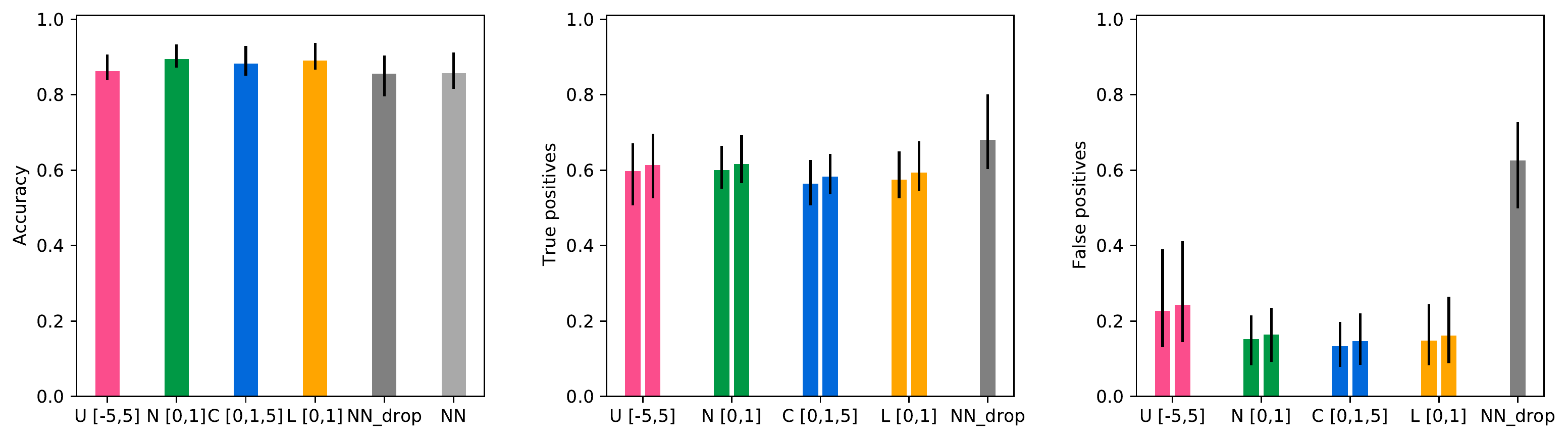}
\caption{Summary of BNN and NN predictions for dataset III, showing the test accuracy of the predictions (a), the true positive rates (b), and false positive rates for out-of-distribution data (c). The results are shown for BNNs with four priors (Uniform, Normal, truncated-Cauchy, and Laplace (color-coded as in Fig. \ref{figPriorPDF}) and an NN with MC dropout (dropout rate = 0.01). For each model we summarized the results of 10 independent runs, trained on a different random subset of 10 of the total 20 classes, and the whiskers show the range of values derived from these 10 subsets.}
\label{simulated_data_stats}
\end{figure}

\begin{figure}[h!]
\centering
\includegraphics[width=1\textwidth]{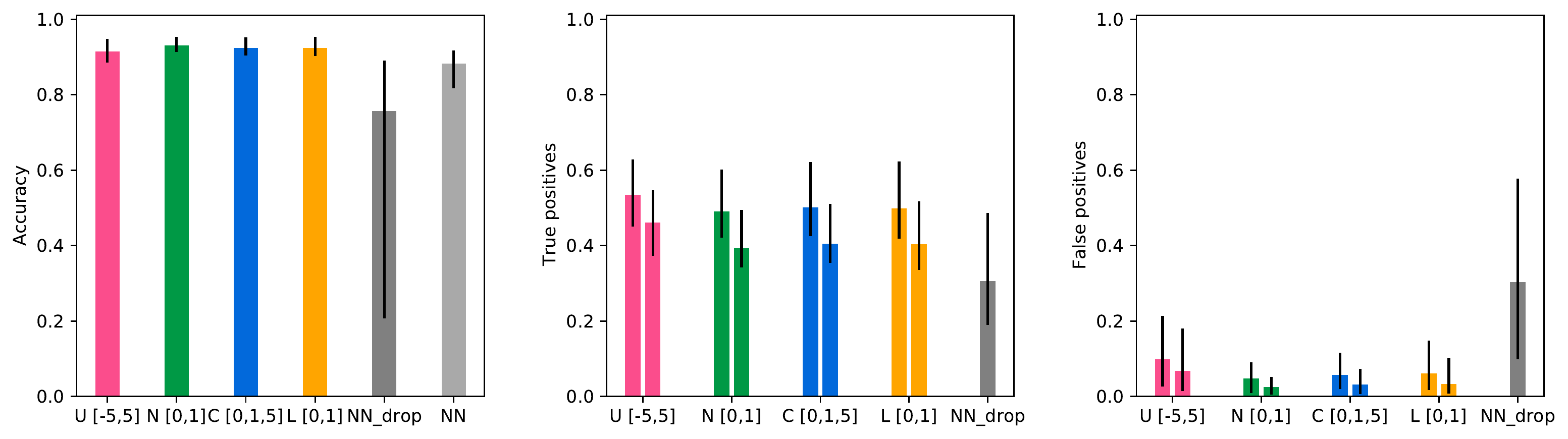}
\caption{Summary of BNN and NN predictions for dataset IV, showing the test accuracy of the predictions (a), the true positive rates (b), and false positive rates for out-of-distribution data (c). The results are shown for BNNs with four priors (Uniform, Normal, truncated-Cauchy, and Laplace (color-coded as in Fig. \ref{figPriorPDF}) and an NN with MC dropout (dropout rate = 0.05). For each model we summarized the results of 10 independent runs, trained on a different random subset of 5 of the total 10 classes, and the whiskers show the range of values derived from these 10 subsets.}
\label{mnist_data_stats}
\end{figure}

The accuracy and informedness obtained for in-distribution data in a simple classification task (dataset I) were greater than 0.95 regardless of the prior (Table \ref{tbl_res}, Fig. \ref{wine_data_predictions}). 
Similar results were obtained for dataset II (Fig. \ref{virus_data_stats}), indicating that the numerical features scored from the two RNA regions considered here can be used to accurately and confidently identify the different Influenza subtypes. 
In dataset III (simulated data) the test accuracy was around 0.85--0.90 and the informedness was around 0.5--0.6, thus showing a substantially lower level of confidence in the predictions (Fig. \ref{simulated_data_stats}). 
The accuracy of dataset IV (a subset of the MNIST data) was greater than 0.90, which is lower then expected \cite{lecun1998mnist} due to the small subset of training instances used here (Fig. \ref{mnist_data_stats}). 
In our tests, the informedness was largely driven by changes in true positive rates, since false positive rates were generally low for in-distribution data.

Our analyses show that different priors can have moderate effects on the test accuracy, which, for datasets III and  IV varied by 3-4\% across prior settings (Table \ref{tbl_res}).
Similarly, the true positive rates showed a variation up to 10\% depending on the dataset and prior. 
For instance in dataset I, a uniform prior yielded a sensitivity of 1, whereas a Laplace prior resulted in a sensitivity of 0.95.
The false positive rates for in-distribution data were only appreciable for datasets III and IV, where the differences associated with prior choice were substantial, despite their limited effect on the overall informedness.
False positive rates under the Laplace prior were about 2-fold lower than under a uniform prior for dataset III and 4-fold lower for dataset IV.

\subsection{Prior choice and out-of-distribution data}
The false positive rates based on out-of-distribution data were significantly higher than for in-distribution data (Table \ref{tbl_res}). 
Prior choice was linked with a variation of false positive rates of ranging from 2-fold (Figs. \ref{simulated_data_stats}, \ref{mnist_data_stats}) to  orders of magnitude (Figs. \ref{virus_data_stats}).
The lowest false positive rates were obtained under different priors depending on the dataset: 
uniform and Laplace for dataset I, Laplace for dataset II, and Cauchy for dataset III, and normal for dataset IV.
The ranking of the models based on informedness for in-distribution data did not match the relative performance of the priors with out-of-distribution data.

\subsection{Bayes factors versus posterior probabilities}
The true and false positive rates obtained from BFs were more conservative than those assessed based on posterior probabilities in dataset I, where both rates were consistently lower (Table \ref{tbl_res}). 
While BF and PP returned comparable results for dataset III, BFs resulted in significantly lower false positive rates for out-of-distribution data for datasets I, II, and IV.

The BFs, as computed here, measure the statistical evidence supporting the best class against all the others combined. However they can also be computed to compare and rank all classes. 
The calculation of BFs explicitly includes the empirical prior probability of each class and for each instance in the dataset, inferred through MCMC.
This might be particularly important when the priors on the weight parameters result in unexpected and unbalanced prior probabilities among classes for a given sample, as often observed in BNNs \cite{wenzel2020}. 
For instance, a prediction for a sample with PP = 0.93 from a model with 5 classes results in moderate BF support if the prior probability is 0.20 (BF = 53), but in very strong support if the prior probability is 0.05, due to unbalanced priors (BF = 252).

\begin{figure}[h!]
\centering
\includegraphics[width=1\textwidth]{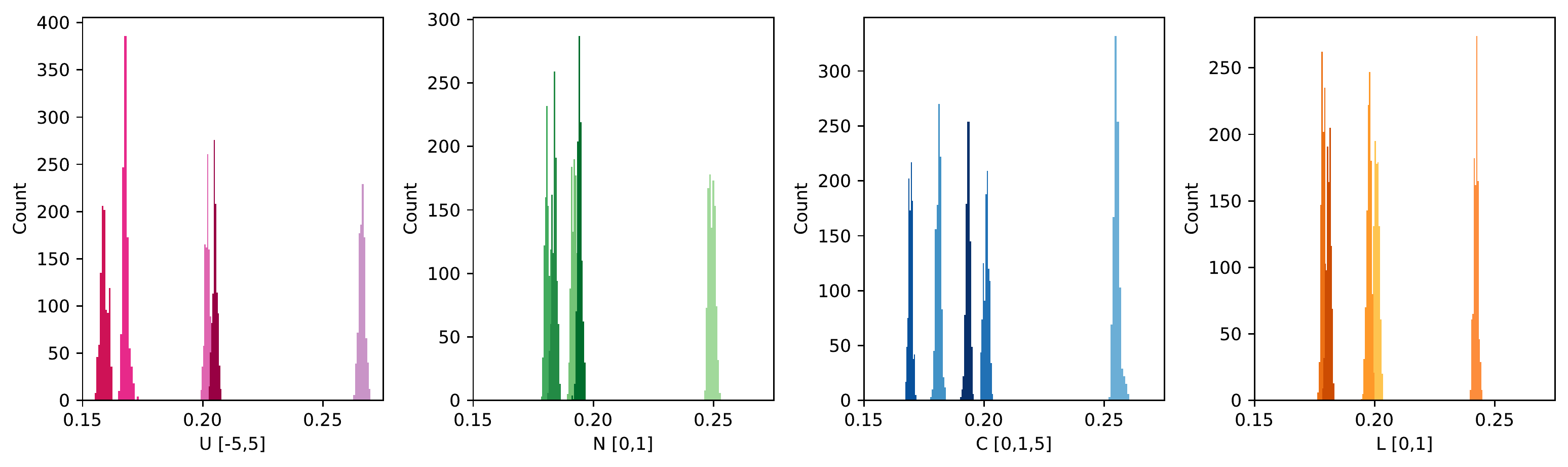}
\caption{Empirical prior probabilities of the 5 training classes across all instances within one of the 10 resampled virus datasets, obtained under four priors: Uniform, Normal, truncated-Cauchy, and Laplace. These plots show that all tested prior distributions applied to the BNN parameters result in unbalanced prior probabilities for the classes, although these probabilities are consistent among the 500 instances within the dataset. To produce these values, we ran an MCMC sampling the weights from their prior distributions. We then used the resulting models to predict labels for each sample (same as Fig. \ref{virus_data_stats}) and plotted the histograms of the resulting label probabilities of each class.}
\label{virus_prior_freq}
\end{figure}

Unbalanced prior probabilities among classes are also observed in our analyses, as shown in Fig. \ref{virus_prior_freq}, and they are not the product of an explicit choice based on prior knowledge, but rather the indirect result of how the features interact with the network architecture and prior distribution on the weight parameters. 
Thus, a measure like BF that uses prior probabilities to quantify the marginal likelihood of a prediction might provide a more informed, if computationally more expensive, way to assess statistical support for a class rather than posterior probabilities.

\subsection{Comparison between BNNs and NNs}
In our tests, BNNs and NNs showed comparable levels of accuracy across the different datasets.
This is in contrast to previous research reporting a lower accuracy of BNNs and showing that a ``cold" MCMC sampler can alleviate this issue (with the caveat that the posterior is no longer exact) \cite{wenzel2020}. 
The close consistency between BNN and NN accuracy in our tests might be linked to the relative small size of the datasets analyzed here and the limited complexity of the networks.

Similarly, the informedness, which combines sensitivity and specificity, did not differ substantially between BNNs and NNs, based on in-distribution data, except for dataset IV, where it was substantially lower based on NN.
However, BNNs strongly outperformed NNs in their ability to identify out-of-distribution data as unknown. The false positive rates of BNNs for out-of-distribution data were orders of magnitude lower for datasets I, II, 3-fold lower for dataset III and 6-fold lower for dataset IV, in comparison with NNs.

Here we used the same network architectures among BNNs and NNs.
It is possible, however, that changing the configuration of the NN model (e.g. number of layers, number of nodes) can bring the performance of NNs closer to that of BNNs \cite{Gal2016, hendrycks2016,lakshminarayanan2016}.

\section{Conclusion}
We tested the effects of using different prior distributions on classification tasks performed through BNNs. 
In our tests we used Metropolis-Hastings MCMC to obtain posterior estimates of the model parameters, but other more efficient algorithms can be used to extend these tests to larger datasets and more complex neural networks \cite{graves2011,hoffman2014nuts, blundell2015, polson2017,zhang2019}.
We show that BFs are an alternative approach to assess the statistical support for the predictions, and hypothesize that they might better account for uneven prior probabilities in classification tasks.  

Understanding the impact of priors on BNNs is important for the application of machine learning methods within a Bayesian framework \cite{lee2004,vladimirova2018}.
While here we focused on small datasets, the sensitivity of BNNs to prior choice for larger datasets might need further exploration. 

Prior choice affects the accuracy of BNNs to some extent, and more strongly the rates of true and false positives.
However, the most significant impact of prior choice is on the ability of a BNN to detect out-of-distributions samples.
False positives in out-of-distribution samples can be interpreted as a measure of the accuracy of a model in distinguishing known samples from unknowns. 
This is important when applying classifiers to datasets extending beyond the described classes used for training the model. 
For instance, with biological data such as the Influenza subtypes used here, the ability to identify unknowns can be a tool to identify samples representing new virus types or undescribed species. 

Gaussian priors are generally the default choice for BNNs  \cite{zhang2017, wenzel2020}, and indeed performed comparatively well in our tests.
However, our results also show that different distributions can yield better results depending on the dataset. 
In the absence of additional knowledge that can inform a decision about the most appropriate priors on the parameters of a neural network, testing and comparing models with different prior distributions, for instance using a validation dataset, can help to substantially improve the performance of classification using BNNs.

\section{Acknowledgements}
We thank Gytis Dudas for advice on the virus data and Stefano Goria and Xavier Meyer for feedback on the manuscript.
D.S. received funding from the Swiss National Science Foundation (PCEFP3\_187012; FN-1749) and from the Swedish Research Council (VR: 2019-04739).

\clearpage


\begin{table}[ht!]
\caption{Accuracy, true positive rate, false positive rate, and informedness for I) the wine dataset, II) the Influenza datasets, and III) the simulated Beta dataset. For BNNs, true and false positive rates are shown based on PP $>$ 0.95 and BF $>$ 150 (in parentheses), whereas for NNs they were based on prediction frequency $>$ 0.95 based on MC dropout. We used the informedness (here indicated with $J$) to rank the performance of the different models. Values reported for datasets II-IV are averaged across 10 replicates based on different partitions of in-distribution and out-of-distribution classes.}
\label{tbl_res}
\begin{center}
\footnotesize
\renewcommand{\arraystretch}{1.1}
\begin{tabular} {l l l l l l l l l}
 \hline
 &  & \multicolumn{4}{c}{In-distribution} & & Out-of-distribution \\
Dataset & Model & Accuracy & True positives & False positives & $J$ & & False positives \\
\hline
I & $\mathcal{U}(-5,5) $     & \textbf{1} & \textbf{1 (0.877)} & 0 (0) & 1     & & \textbf{0.042} (0)  \\ 
 & $\mathcal{N}(0,1)  $      & \textbf{1} & 0.954 (0.785) & 0 (0) & 0.954 & & 0.063 (0)  \\ 
 & $\mathcal{C}_T(0,1,5)$    & \textbf{1} & 0.985 (0.823) & 0 (0) & 0.985 & & 0.083 (0)  \\ 
 & $\mathcal{L}(0,1)  $      & \textbf{1} & 0.954 (0.769) & 0 (0) & 0.954 & & 0.042 (0)  \\ 
 & NN (dropout)              & \textbf{1} & \textbf{1}    & 0     & 0.946 & & 0.313      \\ 
 & NN                        & \textbf{1} & -         & -     & - & & -      \\
\hline
II & $\mathcal{U}(-5,5) $    & \textbf{1}     & 0.989 (0.983) & 0 (0) & 0.989 & & 0.014     (0.005)    \\ 
 & $\mathcal{N}(0,1)  $      & \textbf{1}     & \textbf{0.997 (0.996)} & 0 (0) & 0.997 & & 0.040     (0.019)    \\ 
 & $\mathcal{C}_T(0,1,5)$    & \textbf{1}     & 0.994 (0.991) & 0 (0) & 0.994 & & 0.002     ($<$0.001) \\ 
 & $\mathcal{L}(0,1)  $      & \textbf{1}     & 0.995 (0.990) & 0 (0) & 0.995 & & \textbf{$<$0.001  ($<$0.001)} \\ 
 & NN (dropout)              & 0.999 & 0.977         & 0     & 0.979 & & 0.332                \\ 
 & NN                        & \textbf{1}     & -         & -     & - & & -      \\ 
\hline
III & $\mathcal{U}(-5,5) $ & 0.863 & 0.598 (0.613) & 0.022 (0.024) & 0.591 & & 0.228 (0.243) \\ 
 & $\mathcal{N}(0,1)     $ & \textbf{0.895} & 0.600 (0.617) & 0.010 (0.011) & 0.576 & & 0.152 (0.164) \\ 
 & $\mathcal{C}_T(0,1,5) $ & 0.883 & 0.564 (0.582) & 0.010 (0.011) & 0.534 & & \textbf{0.133 (0.147)} \\ 
 & $\mathcal{L}(0,1)     $ & 0.890 & 0.575 (0.593) & 0.009 (0.010) & 0.550 & & 0.148 (0.161) \\ 
 & NN (dropout)            & 0.856 & \textbf{0.681}         & 0.061         & 0.620 & & 0.626         \\ 
 & NN                      & 0.857     & -         & -     & - & & -      \\
\hline
IV  & $\mathcal{U}(-5,5) $ & 0.914 & \textbf{0.535 (0.462)} & 0.003    (0.001)    & 0.769 & & 0.098 (0.068) \\ 
 & $\mathcal{N}(0,1)     $ & \textbf{0.931} & 0.490 (0.394) & $<$0.001 ($<$0.001) & 0.736 & & \textbf{0.047 (0.025)} \\ 
 & $\mathcal{C}_T(0,1,5) $ & 0.923 & 0.501 (0.406) & 0.001    ($<$0.001) & 0.752 & & 0.057 (0.032) \\ 
 & $\mathcal{L}(0,1)     $ & 0.924 & 0.499 (0.403) & 0.001    ($<$0.001) & 0.753 & & 0.061 (0.034) \\ 
 & NN (dropout)            & 0.757 & 0.306         & 0.079         & 0.227 & & 0.303         \\ 
 & NN                      & 0.882     & -         & -     & - & & -      \\ 
\hline& 
\end{tabular}
\end{center}
\end{table}


\clearpage

\clearpage

\begin{thebibliography}{9}
  
  
\bibitem{amodei2016}
D. Amodei, C. Olah, J. Steinhardt, P. Christiano, J. Schulman, and D. Mane.  \textit{Concrete problems in ai safety.} arXiv: 1606.06565, 2016.

\bibitem{bao2017}
W. Bao, J. Yue, and Y. Rao. \textit{A deep learning framework for financial time series using stacked autoencoders and long-short term memory}. PloS one, 12(7), 2017.

\bibitem{blundell2015} 
C. Blundell, J. Cornebise, K. Kavukcuoglu, and D. Wierstra. \textit{Weight uncertainty in neural networks.} arXiv: 1505.05424, 2015.

\bibitem{Gal2016}
Y. Gal and Z. Ghahramani. \textit{Dropout as a Bayesian approximation: Representing model uncertainty in deep learning. } Proceedings of the 33rd International Conference on Machine Learning: 1612.01474, 2016.

\bibitem{Gelman_et_al2013}
A. Gelman, J. B. Carlin, H. S. Stern, and D. B. Rubin. \textit{Bayesian Data Analysis}, Third Edition (Chapman \& Hall/CRC Texts in Statistical Science). 2013.

\bibitem{Goodfellow2016}
I. Goodfellow, Y. Bengio, and A. Courville. \textit{Deep Learning}. MIT Press, 2016. http: //www.deeplearningbook.org.

\bibitem{graves2011}
A. Graves. \textit{Practical variational inference for neural networks.} In Advances in neural information processing systems, pages 2348–2356, 2011.

\bibitem{hendrycks2016}
D. Hendrycks and K. Gimpel. \textit{A baseline for detecting misclassified and out-of- distribution examples in neural networks}. arXiv, page 1610.02136, 2016.

\bibitem{hoffman2014nuts}
M. D. Hoffman and A. Gelman. \textit{The no-u-turn sampler: adaptively setting path lengths in Hamiltonian Monte Carlo}. Journal of Machine Learning Research, 15(1):1593–1623, 2014.

\bibitem{Kass_1995}
R. E. Kass and A. E. Raftery. \textit{Bayes Factors}. Journal of the American Statistical Association, 90(430):773–795, 1995.

\bibitem{kingma2014adam}
D. P. Kingma and J. Ba. \textit{Adam: A method for stochastic optimization}. arXiv, page 1412.6980, 2014.

\bibitem{lakshminarayanan2016}
B. Lakshminarayanan, A. Pritzel, and C. Blundell. \textit{Simple and scalable predictive uncertainty estimation using deep ensembles},  Advances in neural information processing systems. 2017.

\bibitem{lecun2015deeplearning}
Y. LeCun, Y. Bengio, and G. Hinton. \textit{Deep learning.} Nature, 521(7553):436–444, 2015.

\bibitem{lecun1998mnist}
LeCun, Yann, et al. \textit{Gradient-based learning applied to document recognition.} Proceedings of the IEEE 86.11 (1998): 2278-2324.

\bibitem{lee2004}
H. K. Lee. \textit{Priors for neural networks}. In Classification, Clustering, and Data Mining Applications, pages 141–150. Springer, 2004.

\bibitem{ReLU2010}
V. Nair and G. E. Hinton. \textit{Rectified linear units improve restricted boltzmann machines.} In Proceedings of the 27th International Conference on International Conference on Machine Learning, ICML10, page 807–814, Madison, WI, USA, 2010. Omnipress.

\bibitem{neal2012}
R. M. Neal. \textit{Bayesian learning for neural networks}, vol. 118. Springer Science \& Business Media, 2012.

\bibitem{pytorch2019}
A. Paszke, S. Gross, F. Massa, A. Lerer, J. Bradbury, G. Chanan, T. Killeen, Z. Lin, N. Gimelshein, L. Antiga, et al. \textit{Pytorch: An imperative style, high-performance deep learning library}. In Advances in Neural Information Processing Systems, pages 8024– 8035, 2019.

\bibitem{polson2017}
N. G. Polson, V. Sokolov, et al. \textit{Deep learning: a Bayesian perspective. } Bayesian Analysis, 12(4):1275–1304, 2017.

\bibitem{powers2011}
D. M. Powers. \textit{Evaluation: from precision, recall and f-measure to roc, informedness, markedness and correlation}. 2011.

\bibitem{severyn2015}
A. Severyn and A. Moschitti. \textit{Twitter sentiment analysis with deep convolutional neural networks.} In Proceedings of the 38th International ACM SIGIR Conference on Research and Development in Information Retrieval, p. 959–962, 2015.

\bibitem{Silvestro2019}
D. Silvestro, S. Castiglione, A. Mondanaro, C. Serio, M. Melchionna, P. Piras, M. Di Febbraro, F. Carotenuto, L. Rook, and P. Raia. \textit{A 450 million years long latitu- dinal gradient in age-dependent extinction}. Ecology Letters, 5:724–8, 2019.

\bibitem{vladimirova2018}
M. Vladimirova, J. Verbeek, P. Mesejo, and J. Arbel. \textit{Understanding priors in bayesian neural networks at the unit level}. arXiv:1810.05193, 2018.

\bibitem{wang2019deep}
M. Wang, L. Yu, D. Zheng, Q. Gan, Y. Gai, Z. Ye, M. Li, J. Zhou, Q. Huang, C. Ma, et al. \textit{Deep graph library: Towards efficient and scalable deep learning on graphs}. arXiv: 1909.01315, 2019.

\bibitem{wang2020}
S. Wang, B. Kang, J. Ma, X. Zeng, M. Xiao, J. Guo, M. Cai, J. Yang, Y. Li, X. Meng, et al. \textit{A deep learning algorithm using ct images to screen for corona virus disease (covid-19)}. medRxiv, 2020.

\bibitem{wenzel2020}
F. Wenzel, K. Roth, B. S. Veeling, J. Swiatkowski, L. Tran, S. Mandt, J. Snoek, T. Salimans, R. Jenatton, and S. Nowozin. \textit{How good is the bayes posterior in deep neural networks really?}, 	arXiv: 2002.02405, 2020.

\bibitem{zhang2017}
G. Zhang, S. Sun, D. Duvenaud, and R. Grosse. \textit{Noisy natural gradient as variational inference},  arXiv: 1712.02390, 2017.

\bibitem{zhang2019}
R. Zhang, C. Li, J. Zhang, C. Chen, and A. G. Wilson.\textit{ Cyclical stochastic gradient mcmc for bayesian deep learning.} arXiv: 1902.03932, 2019.

\bibitem{zhang2017influenza}
Y. Zhang, B. D. Aevermann, T. K. Anderson, D. F. Burke, G. Dauphin, Z. Gu, S. He, S. Kumar, C. N. Larsen, A. J. Lee, et al. \textit{Influenza research database: an integrated bioinformatics resource for influenza virus research.} Nucleic acids research, 45(D1):D466–D474, 2017.

\bibitem{zhou2015}
J. Zhou and O. G. Troyanskaya. \textit{Predicting effects of noncoding variants with deep learning–based sequence model.} Nature methods, 12(10):931–934, 2015.
%
%
%
%
%
%
%



\end{thebibliography}

\bibliographystyle{unsrt}

\end{document}